# Urban Change Detection by Fully Convolutional Siamese Concatenate Network with Attention

Farnoosh Heidary[a], Mehran Yazdi[b], Maryam Dehghani[a], and Peyman Setoodeh[b]

[a]Department of Civil and Environmental Engineering, School of Engineering, Shiraz University, Shiraz, Iran.

[b]School of Electrical and Computer Engineering, Shiraz University, Shiraz, Iran.

ABSTRACT

Change detection (CD) is an important problem in remote sensing, especially in disaster time for urban management. Most existing traditional methods for change detection are categorized based on pixel or objects. Object-based models are preferred to pixel-based methods for handling very high-resolution remote sensing (VHR RS) images. Such methods can benefit from the ongoing research on deep learning. In this paper, a fully automatic change-detection algorithm on VHR RS images is proposed that deploys Fully Convolutional Siamese Concatenate networks (FC-Siam-Conc). The proposed method uses preprocessing and an attention gate layer to improve accuracy. Gaussian attention (GA) as a soft visual attention mechanism is used for preprocessing. GA helps the network to handle feature maps like biological visual systems. Since the GA parameters cannot be adjusted during network training, an attention gate layer is introduced to play the role of GA with parameters that can be tuned among other network parameters. Experimental results obtained on Onera Satellite Change Detection (OSCD) and RIVER-CD datasets confirm the superiority of the proposed architecture over the state-of-the-art algorithms.

**Introduction**

Remote sensing is the technology of obtaining information about an object/region without physically contacting the object/region in question, using satellites, airplanes, or drones. One of the main purposes of remote sensing is to observe the evolution of the earth. Satellite and aerial imaging systems allow us to track changes that occur around the world, both in densely populated areas and in remote areas that are difficult to access. Increase in the number of satellites has facilitated the acquisition of remote sensing images as well as easy and free access to these images (Bruzzone and Bovolo, 2013). This has paved the way for working on various applications such as atmospheric parameter modeling, change detection, terrain monitoring, etc. (Rasp, Pritchard and Gentine, 2018)(Sidike *et al.*, 2019)(De Chant and Kelly, 2009). Change detection is one of the most important problems in remote sensing and is essential for accurate analysis of the earth's surface in a period of time(Lu *et al.*, 2004).

Change detection means identifying areas of the earth's surface that have changed over time from two or more images taken from a scene in different times(Lu *et al.*, 2004). These images should be coregistered before any further analysis. Changes in the earth's surface may be caused by natural disasters, urban development, deforestation etc. More accurate detection of the changes results in better analysis of land evolution over a period. Spectral changes are commonly used to detect distance measurement changes in the time series images, which may be occurred in their shape and texture or spectral characteristics(Bruzzone *et al.*, 2000). Detecting changes manually is a slow and laborious process (Singh, 1989). On the other hand, the automatic changes detection problem by using image pairs or sequences has been studied for decades. The algorithms history for general change detection and an extensive review of methods are brought in (Singh, 1989) and (Hussain *et al.*, 2013) respectively. One of the main problems with these methods is their results dependency on the algorithm's compatibility with the input data, meaning that no algorithm has the same accuracy for different inputs. Hereby, deep neural networks have been recently used for change detection due to their capability to be trained by applying different input data and response to a vast variation of inputs (Vignesh, Thyagharajan and Ramya, 2019).

As noted by Hussain et al (Hussain *et al.*, 2013), change detection methods can be roughly divided into two categories: pixel based and object/region based methods. The first method attempts to determine whether there is a change in each pixel in image pairs, while the second method uses neighbourhood information such as colour, shape and texture to divide the image into objects/regions and then detect changes in them. Change detection algorithms can also be divided into supervised and unsupervised learning. Due to vast available annotated datasets lack for change detection, unsupervised methods are more favourable (Hussain *et al.*, 2013)(Vakalopoulou *et al.*, 2015)(Sidike *et al.*, 2019). Many of these methods automatically analyze different images and identify appropriate patterns useful for change detection (Bazi, Bruzzone and Melgani, 2004)(Bruzzone and Prieto, 2000). Some others use unsupervised learning methods such as repetitive training (Caye Daudt *et al.*, 2019), automatic encoding (Maoguo Gong *et al.*, 2013), and K-means clustering (Celik, 2009) to separate changed pixels from those that have not been changed, however their success is limited. Meanwhile, methods for detecting changes have been also proposed using supervised learning algorithms such as support vector machines (Bruzzone and Bovolo, 2013) (Caye Daudt *et al.*, 2019) random forests (Sesnie *et al.*, 2008), and neural networks (Gopal, Woodcock and Member, 1996)(Dai and Khorram, 1999), but feature selection strategy is a main challenge in these algorithms and their success is also limited. Recently, Convolutional Neural Networks (CNN) have been introduced for the supervised change detection (Zhan *et al.*, 2017). Most of these methods use data transfer techniques, in which a network already trained for another purpose on a large dataset, is used to prevent data shortage problems. For instance, Mohamad El Amin et al. used the trained network parameters to extract feature maps and create the difference image, and then sharpen the pixels that were changed (Mohammed El Amin, Liu and Wang, 2016). While transfer learning can be a valid solution, it has also limitations, The transfer learning assumes that all images are of the same type, but it cannot be true for some cases such as multi-spectral images where using all bands leads to better results in change detection (Caye Daudt *et al.*, 2019).

In term of input data for change detection, images can be optical, radar or multispectral where based on input data the algorithms may be different (Dreschler-Fischer *et al.*, 1993)(Song and Woodcock, 2003). Change detection in optical images is usually performed based on the difference in pixel intensities, and in the case of multispectral images, it is performed based on differences in spectral bands. For instance, pixel-to-pixel methods can be used to obtain the difference image in optical images (Bruzzone and Bovolo, 2013)(Mohammed El Amin, Liu and Wang, 2016). However, the accuracy can be improved by using the region based methods by segmenting images into different regions and then by coregistering images, and taking into account the difference in similar regions in two images, the change detection is achieved (Kaur and Goyal, 2015). Another way to get a difference image is through using wavelet transforms of images and working with the wavelet coefficients of the two images. In SAR images, the change detection methods are based on the two images pixel intensity ratio due to the multiplying nature of the sensor noise (Howarth and Wick Ware, 1981). In this type of images, the difference image usually comes from the difference of the two images logarithm (Prendes *et al.*, 2015).

When annotated datasets are available, supervised classifiers show a better performance than unsupervised classifiers ( El Amin et al., 2018)(Bazi, Bruzzone and Melgani, 2004). In our work, we also use deep learning networks as a supervised segmentation by introducing two images taken from a scene in different time and expect that the trained network, by automatically producing optimal features, segmented image pixels into changes and unchanged classes.

In this paper, we propose specific CNN architecture for tackling change detection. Our proposed architecture benefits from the combination of U-Net, Siamese model, and attention gate to extract feature maps from VHR images in a supervised manner. In this architecture, two methods are used for improving results; preprocessing of the dataset by gaussian attention and using attention gate for up sampling in the network decoder part.

The paper is organized as follows. Section 2 describes in detail the proposed architecture of attention-based Siamese network for change detection. Section 3 contains quantitative and qualitative comparisons with previous change detection methods. Finally, Section 4 contains the concluding remarks.

**Methodology**

The change detection task can be treated as a problem of binary image segmentation, it is common to introduce advanced semantic segmentation architectures to solve it (Zhang *et al.*, 2018)(Aich, Van Der Kamp and Stavness, 2018).The proposed algorithm for performing change detection on VHR images, was inspired by fully convolutional Siamese concatenate (Caye Daudt, Le Saux and Boulch, 2018), U-Net with attention gate layer (Oktay *et al.*, 2018), and Siamese ( El Amin et al., 2018) (FC-Siam-conc-Att), all developed as end-to-end architectures. The proposed method is illustrated in Fig. 1. Two coregistered images, after applying Gaussian Attention (GA) (Adam Kosiorek, 2017), are fed as inputs to the Siamese network. The model learns multiscale representations and different semantic levels of visual features. We call the network that learns with these preprocess dataset FC-Siam-conc-Att-GA. After extracting feature maps, skip connections are adopted to use feature maps in the decoder part. Finally, this subnetwork is followed by a sigmoid layer to generate the final change map.

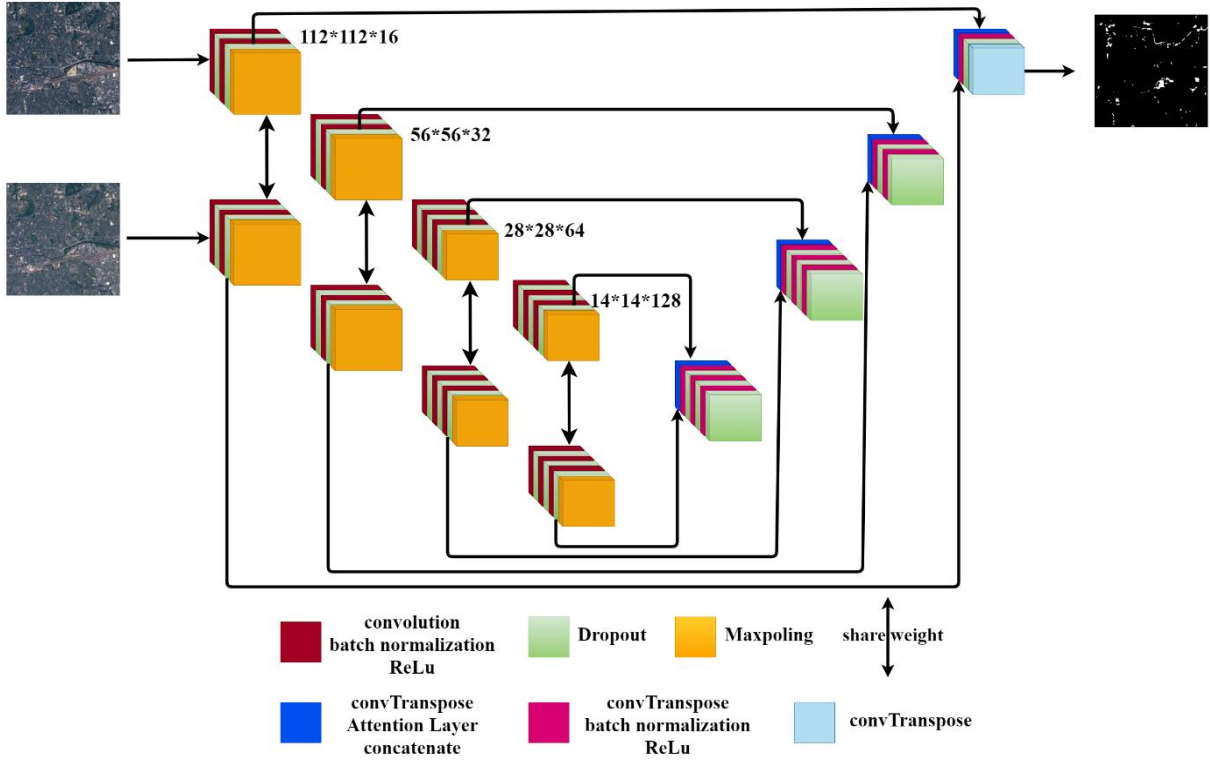

Figure 1: Aarchitecture of FC-siam-conc-Att network.

*Preprocessing the dataset*

GA helps the network to see feature maps as human does (Adam Kosiorek, 2017), but its parameters cannot be trained on the network. GA uses a Gaussian filter to create attention maps with the same size as original images. To use GA, at the first Glimpse should be define. So, you can write Glimpse using the equivalent of (1) (Vakalopoulou, M. *et al.,* 2015).

$$g = A_y I A_x^T \qquad (1)$$

Such filters can be mathematically represented by matrices $A_x \in R^{w*W}$ and $A_y \in R^{h*H}$ for an image with dimensions of H * W (where w,h are the small part of image). A Glimpse of the image is denoted by $g$, which is the result of applying attention to image *I*. Each one of these matrices has a Gaussian parameter in each row, and the distance (per column unit) between the Gaussian centers in successive rows is specified by the parameter *d*. After defining Glimpse, $A_y$ is define as an equivalent of (2) (Vakalopoulou, M. *et al.,* 2015).

$$A_y = Gussian\_mask(u, s, d, h, H) \quad (2)$$

The Gaussian mask is characterized by the parameters u, s, d, R, and C, where *u* denotes the first gaussian center, *s* is the Gaussian standard deviation, *d* is the distance between Gaussian centers, *R* is the rows number in the mask (there is one Gaussian per line), and *C* is the columns number in the mask (Vakalopoulou, M. *et al.*, 2015).

*Model architecture*

U-Net with the attention gate layer has great benefits for extracting multiscale feature maps using multi-level convolution pathways. Attention allows to focus on locations that changes have occurred. In deep networks, attention gates can be used in different parts of the network, such as in the feature maps or in the decoder before the feature maps are concatenated together or in the network output. In this paper, the proposed attention gate in (Ronneberger, Fischer and Brox, 2015) has been used. This gateway automatically learns to focus on target structures of different shapes and sizes. Gate-trained models implicitly learn to ignore irrelevant areas in an input image while highlighting specific features. The attention gateway can be easily integrated into the CNN-based architectures such as the U-Net model with minimal computational load while increasing the model sensitivity and predictive accuracy. Experimental results on different datasets with different image sizes show that the focus gateway improves the U-Net predictive performance while maintaining computational efficiency (Oktay *et al.*, 2018). The most significant difference between U-Net and the proposed network is in the decoder part as shown in Figure 2. Siamese network was used in the encoder and attention gate layer was used in the decoder.

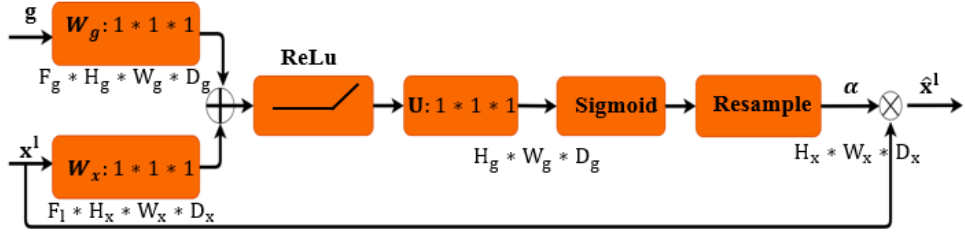

Figure 2 Attention gate layer (Oktay *et al.*, 2018).

*Loss function*

Our proposed architecture performs an imbalanced binary segmentation. Since the problem is an imbalanced task, for improving results, we used a weighted binary cross-entropy and weighted dice cross-entropy combination as follows:

$$E = E_{wce} + E_{wdice} \qquad (3)$$

where $E_{wce}$ refers to the weighted cross entropy and $E_{wdice}$ is the weighted dice coefficient loss.

*Weighted cross-entropy*

In the weighted cross-entropy, weights are assigned to all positive samples to compensate for the imbalanced datasets. For example, in our problem, the number of white pixels is more than the number of black pixels, balanced binary cross-entropy does not work very well. The weighted cross-entropy is calculated as (Jadon, 2020):

$$E_{wce} = -(\beta p log(\hat{p}) + (1-p)\log(1-\hat{p})) \qquad (4)$$

where $p$ is the ground truth, $\hat{p}$ is the prediction, and $\beta$ is a coefficient for balanced classes.

*Weighted dice coefficient loss*

Weighted dice coefficient loss is a distance measure between two samples. This measure takes values between 0 and 1, where 1 means that the two samples are the same. Weighted dice coefficient loss is calculated as (Jadon, 2020):

$$E_{wdice} = 1 - \frac{2\beta p\hat{p}+1}{\beta p+\beta\hat{p}+1} \quad (5)$$

where $p$ is the ground truth, $\hat{p}$ is the prediction, and $\beta$ is a coefficient for the balanced classes.

*Metrics*

To evaluate the proposed network accuracy, four metrics, i.e. recall, F1, accuracy and precision has been employed.

$$Recall = \frac{TP}{TP + FN} \quad (6)$$

$$F1 = 2 * \frac{precision * Recall}{precision + Recall} \quad (7)$$

$$Accuracy = \frac{TP + TN}{TP + FP + TN + FN} \quad (8)$$

$$precision = \frac{TP}{TP + FP} \quad (9)$$

where TP, FP, TN, and FN denote the number of true positive, false positive, true negative, and false negative classification results, respectively. When the precision is high, it means that the number of false alarms is small, and when the recall is high, it means that the number of misdetections is small.

**Experimental results**

We have evaluated and compared our proposed network on OSCD and LIVER-CD datasets. In this section, these datasets are described in detail, then the parameters setting is brought and finally, the experimental results are presented.

*Dataset and evaluation metrics*

The Onera Satellite Change Detection (OSCD) dataset is used to train and test the proposed network. This dataset contains Sentinel 2 multidimensional images acquired between 2015 and 2018 from various locations selected around the world including Brazil, USA, Europe, Middle East, and Asia. For each location, image pairs containing 13 bands of multidimensional satellite images were produced by the Sentinel 2 satellite. The spatial resolution of the images varies between 10, 20 and 60 meters. A pixel-level change ground truth is provided for 14 image pairs. In this dataset, change detection mainly focuses on detecting urban changes such as detecting changes due to new buildings and new streets. This dataset can be used to train and adjust the parameters of the change detection algorithms (Caye Daudt, Le Saux and Boulch, 2018).

In addition, LIVER-CD dataset has been used in our work. The dataset was recently released with 637 high-resolution images. This dataset includes different types of buildings, such as villa residences, tall apartments, small garages and large warehouses. This dataset mainly focuses on building-related changes, uch as mended structures, a decrease in the number of buildings and an increase in that i.e. turning soil, grass, hardened ground under construction-building turned to a new finished region and and the building decline. These bitemporal images are annotated by visual interpretations carried out by remote sensing experts using binary labels (1 for changed and 0 for unchanged pixels) (Chen and Shi, 2020).

*Parameter setting and results*

The first step to implement the algorithm for change detection is to define the Gaussian parameters (i.e., u, s, d, R, and C applied on training images). Since these parameters are not learned by the network, we should find the best values for these parameters for which the network achieves better results by trial and error. For training the network, dataset was first preprocessed. Results of training the network on preprocessed OSCD with different hyper-parameters are shown in Table 1. Size of the convolution kernel is set to 3×3 for all convolution layers. The number of convolutional filters in the encoder part is set to [16, 32, 64, 128] and the number of convolutional filters in the decoder part is set to [128,64,32,16] . Two RGB images with the size of 112×112 are fed into the network. Input is an image with 112×112×3 pixels, while output is a tensor with 112×112×1 pixels.

Table 1 Results obtained by training different networks and setting different hyper-parameters.

| network | u | s | d | Recall | F1 | Precision | accuracy |
|---|---|---|---|---|---|---|---|
| FC-Siam-conc | 0.1 | 0.5 | 2 | 69.92 | 64.86 | 62.25 | 91.04 |
| | 0.1 | 0.5 | 1 | 70.64 | 66.49 | 64 | 91.97 |
| | 0.1 | 0.5 | 5 | 70.12 | 65.21 | 62.6 | 91.23 |
| FC-Siam-conc-Att | 0.1 | 0.5 | 2 | 71.47 | 68.29 | 66.08 | 92.83 |
| | 0.1 | 0.5 | 1 | **75.57** | 68.26 | 64.75 | 91.56 |
| | 0.1 | 0.5 | 5 | 71.64 | 67.67 | 65.16 | 92.4 |
| FC-Siam-diff | 0.1 | 0.5 | 2 | 73.95 | 67.13 | 63.86 | 91.28 |
| | 0.1 | 0.5 | 1 | 73.04 | 65.93 | 62.77 | 90.7 |
| | 0.1 | 0.5 | 5 | 69.31 | 66.32 | 64.29 | 92.33 |
| FC-Siam-diff-Att | 0.1 | 0.5 | 2 | 71.19 | **69.31** | **67.8** | **93.5** |
| | 0.1 | 0.5 | 1 | 71.71 | 67.69 | 65.86 | 92.39 |
| | 0.1 | 0.5 | 5 | 74.25 | 67.93 | 64.67 | 91.73 |

As shown in Table 1, the four metrics of F1, precision, recall, and accuracy are used to evaluate the network performance. While FC-Siam-diff-Att provided better results in terms of precision, F1, and accuracy, FC-Siam-conc-Att performed better in terms of recall. Therefore, it can be concluded that using the attention gate improves the results. However, by reducing the pixel-by-pixel feature maps, the values of the pixels may be reduced too much, and some pixels may not be the focus of attention. Except for FC-Siam-conc, in the other 3 networks, d=2 led to the best results. It can be concluded that, as the distance between Gaussian centers decreases, accuracy of the network decreases to the information integration. On the other hand, increase in the distance between Gaussians centers leads to the loss of full attention on the image. For FC-Siam-conc, due to the lack of attention and differentiation of feature maps, decreasing $d$ has led to the examination of more pixel values of input images, thus, improves the network accuracy. In order to verify the effectiveness and the superiority of the proposed change detection method, results on a test image, which consists of a complex area, are presented in Figure 3.

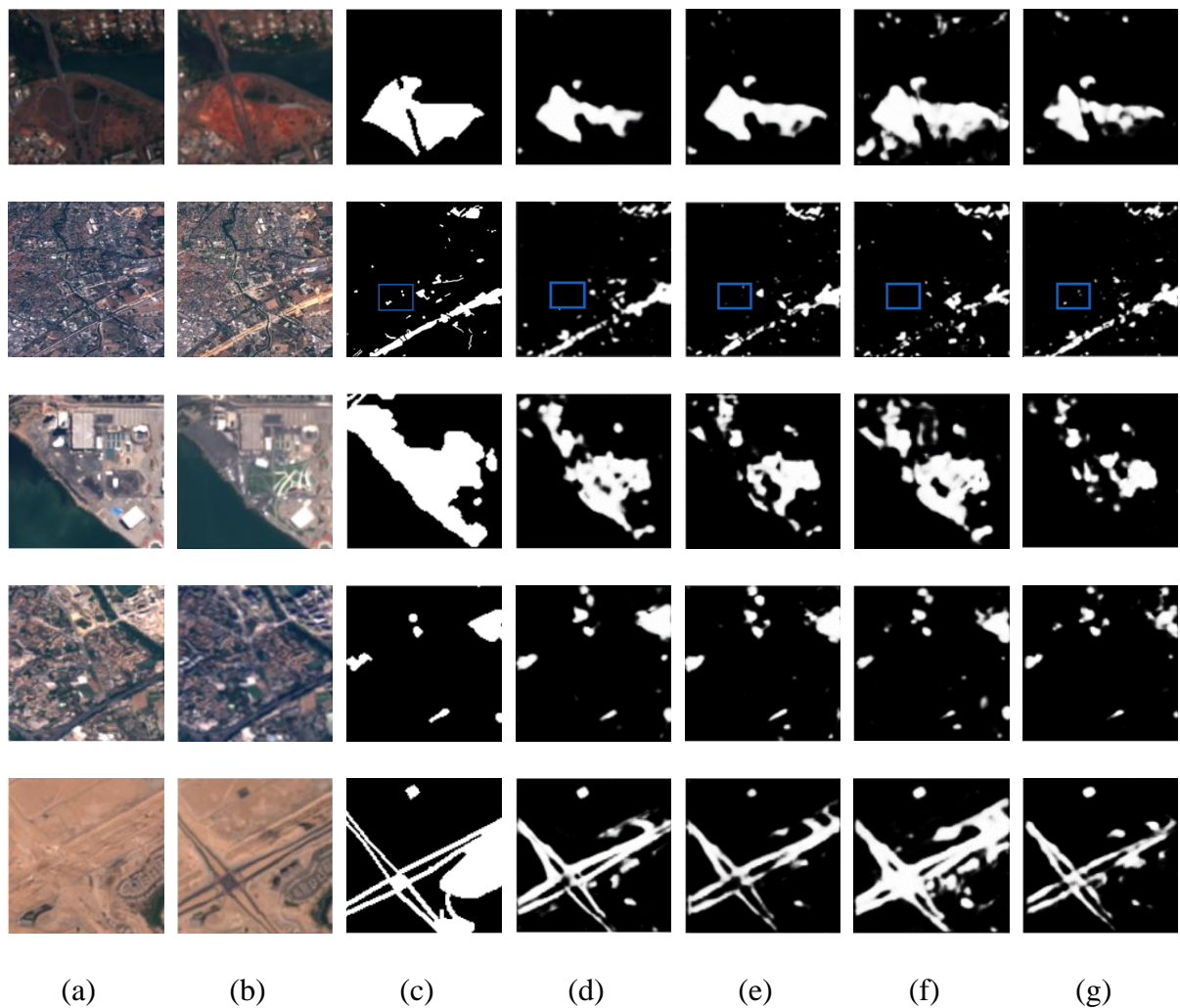

(a)  (b)  (c)  (d)  (e)  (f)  (g)

Figure 3 Visual comparison of the change detection results using different approaches on the preprocessed OSCD dataset: (a) Image T1, (b) Image T2, (c) Ground Truth, (d) FC-Siam-conc (d=1), (e) FC-Siam-conc-Att (d=2), (f) FC-Siam-diff (d=2), (g) FC-Siam-diff-Att (d=2).

As shown, the proposed FC-Siam-diff-Att architecture provides more details compared to other methods. Qualitatively speaking, among the four architectures, the proposed FC-Siam-diff-Att method provides the closest result to the ground truth.

Table 2 compares the previously mentioned four architectures performance, which were trained on the OSCD dataset with the best hyper-parameter values presented in Table 1, with other networks.

Table 2  Training results of applied networks on OSCD dataset.

| Network | F1 | Precision | Recall | Accuracy |
|---|---|---|---|---|
| **FC-Siam-conc** | 64.26 | 61.95 | 68.42 | 91.19 |
| **FC-Siam-diff** | 65.88 | 63.91 | 68.78 | 92.24 |
| **FC-Siam-conc-Att** | 67.96 | 61.95 | 73.07 | 92.13 |
| **FC-Siam-diff-Att** | **69.39** | **67.47** | 71.86 | **93.33** |
| **CDNet** | 66.97 | 64.16 | 71.99 | 91.82 |
| **Unet** | 68.83 | 65.79 | **74.01** | 92.37 |
| **Unet++** | 68.93 | 66.41 | 72.74 | 92.82 |
| **Unet++KSOF** | 67.28 | 64.69 | 71.51 | 92.19 |
| **WVnet** | 65.65 | 62.79 | 71.4 | 91.08 |
| **STANet** | 62.55 | 60.53 | 66.26 | 90.76 |

Generally speaking, the best results were obtained by those networks, which deployed the attention gate. It can be concluded from Table 2 that the attention gate application improves the algorithm performance. The visual test results for these networks are illustrated in Figure 4.

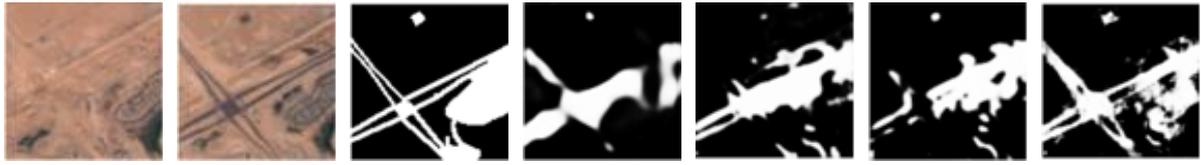

(a) (b) (c) (d) (e) (f) (g)

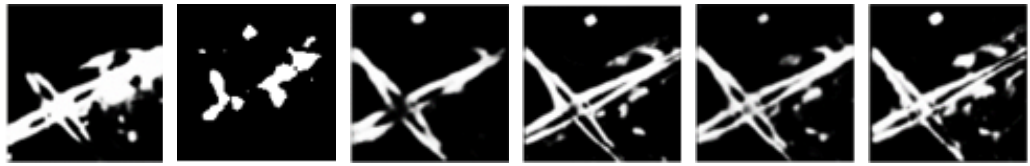

(h) (i) (j) (k) (l) (m)

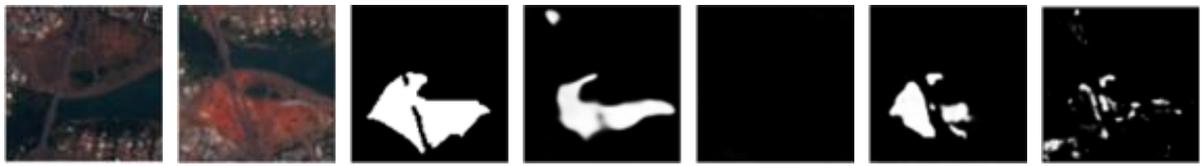

(a) (b) (c) (d) (e) (f) (g)

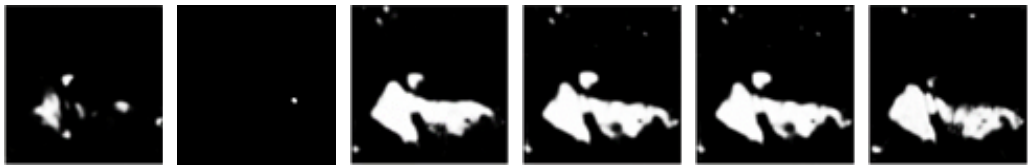

(h) (i) (j) (k) (l) (m)

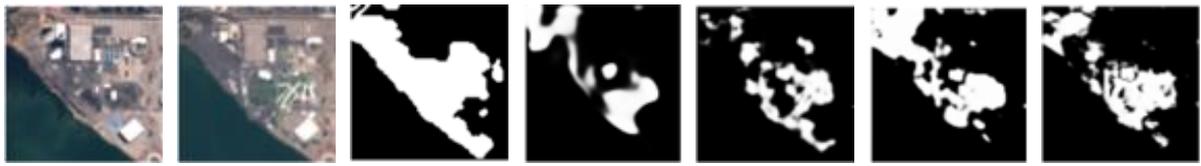

(a) (b) (c) (d) (e) (f) (g)

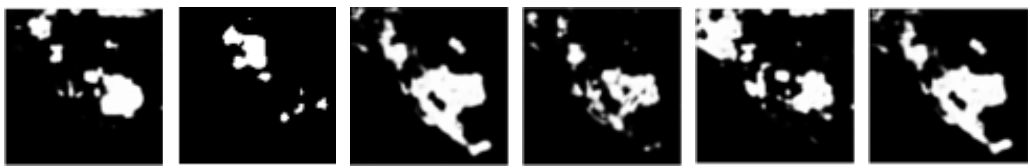

(h) (i) (j) (k) (l) (m)

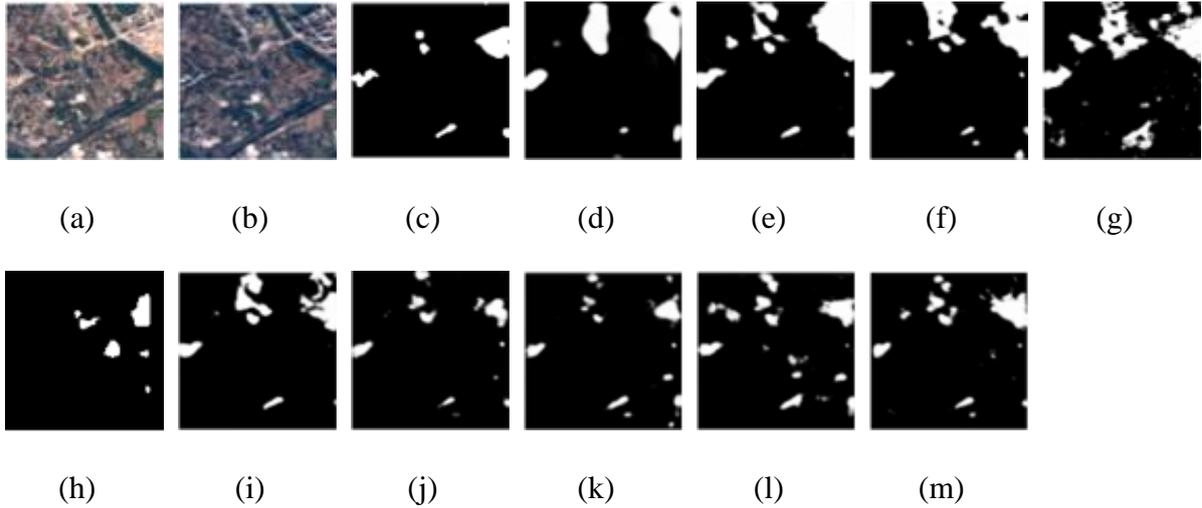

(a) (b) (c) (d) (e) (f) (g)

(h) (i) (j) (k) (l) (m)

Figure 4 Visual comparisons of change detection results using different approaches: (a) image T1, (b) image T2, (c) Ground Trouth, (d) CDNet, (e) UNet, (f) UNet++, (g) Unet++KSOF, (h) WVnet, (i) STANet, (j) FC-Siam-diff-Att, (k) FC-Siam-conc, (l) FC-Siam-conc-Att, (m) FC-Siam-diff.

As observed, one of the closest predictions to the ground truth is the prediction made by FC-Siam-diff-att. Here, the use of attention has led to more accurate predictions as well.

To further evaluate the proposed network performance, its performance was examined on a new and simple dataset called RIVER-CD. Table 3 compares the performance of the proposed network trained on this dataset with those of three other networks presented in (Chen and Shi, 2020).

Table 3 Results achieved by networks different trained on the RIVER-CD dataset.

| Network | F1 | Precision | Recall |
|---|---|---|---|
| **FC-Siam-diff-Att-GA** | **88.46** | **90.56** | 86.6 |
| **STAnet (BASE) [18]** | 83.9 | 79.2 | 89.1 |
| **STAnet (BAM) [18]** | 85.7 | 81.5 | 90.4 |
| **STAnet (PAM) [18]** | 87.3 | 83.8 | 91 |

The RIVER-CD dataset has a resolution of 0.5 meters, which is much higher than the resolution of 10, 20, and 30 meters for OSCD. Change detection accuracy can be improved by using images with higher resolutions. The image size of this dataset is 1024 × 1024. To import these images into the proposed network, the size of images was reduced to 256 × 256. Two RGB images with the size of 256×256 are fed into the network. Input is an image with 256×256×3 pixels, while output is a tensor with 256×256×1 pixels.

On this dataset, proposed FC-Siam-diff-Att-GA performance was better than the best network presented in (Lu, D. *et al.*, 2004) by about 7%. The F1 measure was about 1% better. However, in the case of the recall metric, as can be observed, STAnet (PAM) performed better. This is due to the decrease in the values of the pixels caused by reducing the feature maps. The network does not take the pixels with small values into account (Adam Kosiorek, 2017).

**Conclusion**

In this paper, an attention gate layer was added to FC-Siam-conc and FC-Siam-diff for end-to-end change detection based on VHR satellite images. In addition, Gaussian attention was used for images preprocessing. Attention was used as a step towards mimicking following the human vision and improving the deep learning architectures performance. Gaussian attention was used for reducing noise and ignoring irrelevant areas. This attention mechanism was also applied to feature maps via an attention gate. Both quantitative and visual results indicate that the accuracy of networks increases by deploying attention. Moreover, a weighted loss function was used to reduce the class imbalance effect. Using the attention gate on FC-Siam-diff network along with preprocessing by Gaussian attention improved F1, precision, and accuracy metrics. However, calculating the difference between feature maps reduces the values assigned to pixels and some of them may be ignored which leads to the recall metric reduction. A mechanism must be investigated to address this issue and compensate for the reduction in recall. This can be considered as future work.